\documentclass[10pt, a4paper]{article}
\usepackage{amsmath}
\usepackage{lrec-coling2024} 
\usepackage{changes}
\usepackage{hyperref}

\title{\textbf{SensoryT5: Infusing Sensorimotor Norms into T5 for Enhanced Fine-grained Emotion Classification}}

\name{Yuhan Xia\textsuperscript{1}, Qingqing Zhao\textsuperscript{2}, Yunfei Long\textsuperscript{1}, Ge Xu\textsuperscript{3}, Jia Wang\textsuperscript{4}} 

\address{\textsuperscript{1}School of Computer Science and Electronic Engineering, University of Essex \\\textsuperscript{2}Institute of Linguistics, Chinese Academy of Social Sciences \\\textsuperscript{3}College of Computer and Control Engineering, Minjiang University \\\textsuperscript{4}Department of Intelligent Science, Xi'an Jiaotong-Liverpool University \\
\{yx23989, yl20051\}@essex.ac.uk\\
         zhaoqq@cass.org.cn,  xuge@pku.edu.cn, Jia.Wang02@xjtlu.edu.cn\\
         }

\abstract{In traditional research approaches, sensory perception and emotion classification have traditionally been considered separate domains. Yet, the significant influence of sensory experiences on emotional responses is undeniable. The natural language processing (NLP) community has often missed the opportunity to merge sensory knowledge with emotion classification. To address this gap, we propose SensoryT5, a neuro-cognitive approach that integrates sensory information into the T5 (Text-to-Text Transfer Transformer) model, designed specifically for fine-grained emotion classification. This methodology incorporates sensory cues into the T5's attention mechanism, enabling a harmonious balance between contextual understanding and sensory awareness. The resulting model amplifies the richness of emotional representations. In rigorous tests across various detailed emotion classification datasets, SensoryT5 showcases improved performance, surpassing both the foundational T5 model and current state-of-the-art works. Notably, SensoryT5's success signifies a pivotal change in the NLP domain, highlighting the potential influence of neuro-cognitive data in refining machine learning models' emotional sensitivity.
 \\ \newline  }

\begin{document}

\maketitleabstract

\section{Introduction}

Affective computing stands at the intersection of technology and human emotions \cite{li2017inferring}, whereby sentiment analysis and emotion recognition  are generally merged to give machines a semblance of human-like emotional understanding. Specifically, sentiment analysis (SA) seeks to decode the attitudes and viewpoints of opinion holders using computational methods \cite{lu2023sentiment}, providing a coarse-grained categories of polarities: positive, negative, or neutral \cite{long2019study}. Driven by recent advancements in deep learning and bolstered by vast labeled datasets, discriminating sentiments in standard contexts has become progressively more tractable. Cutting-edge models, including the likes of BERT \cite{devlin2018bert}, XLNet \cite{yang2019xlnet}, and the T5 \cite{raffel2020exploring} series, have consistently set benchmarks, achieving high accuracies on an array of sentiment classification tasks.

By contrast, emotion analysis (EA) has received less notable results in recent years. One of the reasons is that different from SA offering a coarse-grained outlook, EA paints a detailed picture. That is, EA not only distinguishes between basic sentiments but also identifies nuanced emotions such as joy, anger, sadness, surprise, and among others \cite{ekman1992argument}. Thus, the task of EA is complicated by the sheer variety of emotional categories. For instance, distinguishing closely related emotions like "contentment" and "happiness" or "annoyance" and "anger" requires a discerning approach, especially when the medium is textual content. Thus, 
this study introduces a SensoryT5 model, tailored to infuse sensory data, which is cognitively more related to emotions and includes linguistically more enriched features, into neural architectures, to achieve a profound comprehension of emotions.

The relationship between emotion and perception/sensation has been verified repeatedly in various disciplines. From a neuroscientific perspective, emotion and sensory information are processed in an overlapping neural region, i.e., the amygdala  \cite{vsimic2021understanding}. Shifting the lens to psychology, emotion and perception are intertwined \cite{zadra2011emotion}. For example, the sense of taste shows an inherent link with reward and aversion mechanisms, such as sucrose being perceived as sweet and desirable, whereas quinine being recognized as bitter and repulsive \citep{yamamoto2008central}. In addition, emotion as a kind of interoception forms an indispensable part of human sensations, when a wide definition of sensory perception adopted \cite{connell2018interoception,lynott2020lancaster}.
In terms of the linguistic conceptualization of emotions, people more frequently use figurative language instead of literal emotion terms  to convey emotions \cite{fainsilber1987metaphorical,lee2018figurative}, and the conceptual metaphor EMOTION IS PERCEPTION is grounded in abundant language usages to show that the human senses are fruitful sources for verbalizing emotions (e.g., sweet and bitter) \cite{lakoff1980metaphors,kovecses2019perception,muller2021metaphorical}.

Given the intertwined relation between emotion and perception/sensation, this study posits that incorporating sensory information into a computational framework can capture the nuanced interplay between them, hence offering a reflection of intricate human affective understanding. Specifically, we utlize the Lancaster Sensorimotor Norms \citep{lynott2020lancaster}, which include language-specific lexical properties representing the correlation between conceptualized lexical meanings and sensory modalities.

Our work boasts three pivotal advancements:
(1) We introduce SensoryT5, an innovative architecture that enhances transformer-based fine-grained emotion classification models by seamlessly embedding sensory knowledge. Marking one of the pioneering endeavors, SensoryT5 is adapted at harmonizing both the nuances of contextual attention and the intricacies of sensory information-based attention.
(2) The SensoryT5 leverages sensorimotor norms within transformer text classification frameworks, contributing to the ongoing efforts to incorporate neuro-cognitive data in NLP tasks. Thus, our work not only demonstrates the practical benefits of this integration in improving emotion classification tasks, but also encourages continued interdisciplinary dialogue and research between the domains of language processing and neuro-cognitive science.
(3) Assessments across multiple real-world datasets pertinent to fine-grained emotion classification affirm that our approach amplifies the efficacy of pre-existing models considerably, even surpassing contemporary state-of-the-art methodologies on selected datasets. This endeavor underscores the value of cognition-anchored data in sculpting attention models. Our findings illuminate the untapped potential of sensory information in refining emotion classification, carving fresh prospects for exploration within the realm of affective computing in NLP.

\section{Related work}
\subsection{Emotion analysis}
Over recent years, the domain of pre-trained language models (PLMs) and large language models (LLMs) has witnessed marked advancements. Noteworthy developments include models like BERT \citep{devlin2018bert}, RoBERTa \citep{liu2019roberta}, GPT-3 \citep{brown2020language}, T5 \citep{raffel2020exploring}, PaLM \citep{chowdhery2022palm}, LLaMA \citep{touvron2023llama} and ChatGPT \citep{openai2023gpt4}. These models, through rigorous pre-training on vast text corpora using self-supervised learning, have the ability to autonomously generate intricate representations. This capability has significantly advanced the field, setting new benchmarks in numerous tasks, notably in sentiment analysis \citep{devlin2018bert,zhang2023sentiment}. An in-depth exploration by \citet{zhang2023sentiment} elucidated the performances of LLMs in sentiment and emotion analysis tasks. The study has highlighted that, while LLMs excel over PLMs in few-shot learning scenarios, PLMs remain superior for more nuanced tasks that demand a deeper understanding of emotions or structured emotional data. Among the discussed models, T5 \citep{raffel2020exploring} stands out due to its innovative 'text-to-text' transfer approach, in which every NLP challenge is remodelled as a text-to-text problem. Consequently, T5 frequently sets the state-of-the-arts in emotion analysis when utilized as the base model.

However, despite the considerable improvements made with these PLMs/LLMs, some research gaps remain relatively fulfilled. Present models, although they possess sophisticated neural architectures capable of discerning patterns from immense text datasets, often overlook the intricate nature of emotion—a dynamic interplay of cognitive and physiological responses triggered by various stimuli \cite{khare2023emotion}. Sensory perceptions, pivotal in shaping these responses, serve as the bedrock upon which our cognitive processes evaluate and generate emotions \cite{niedenthal2019does}. Integrating these models with sensory data can potentially elevate their performance, nudging them closer to approaching human-like comprehension. This presents a significant research opportunity: equipping already potent PLMs/LLMs with an element of sensory perception, an aspect they conventionally lack. With our proposed SensoryT5 model, our ambition is to fill this gap by synergizing the strengths of T5 and augmenting it with sensory knowledge, thereby enabling a deeper and more nuanced understanding of emotions.
\subsection{Cognition-grounded resources: Sensorimotor norms}

In recent years, there is an emergent trend that neuro-cognitive data and computational approaches are synergized in NLP studies. This interdisciplinary synergy unlocks new dimensions in understanding language, sentiment, and emotion, reflecting more accurately the human experience and mental processing. For instance, \citet{long2019study} improved the attention model for sentiment analysis by incorporating a eye-tracking dataset. \citet{chen2021metaphor} incorporated brain measurement data for modeling word embedding. \citet{wan2023perceptional} demonstrated the superiority of neural networks for metaphor detection by leveraging sensorimotor knowledge. These studies collectively underscore a broader shift in the field towards a more integrated approach to NLP. By weaving in neuro-cognitive data, researchers are equipping computational models with a richer and more intricate understanding of human language and cognition, which are often overlooked by traditional data-driven methods.

Given the intimate connection between emotion and perception as demonstrated in various studies reviewed in the last section, this study assumes that a cognitively and linguistically motivated representation of words in text based on sensorimotor knowledge would improve the performance of computational models for emotion analysis. That is not only because sensory inputs are crucial sources of emotions, but also because emotional responses are part of sensory perceptions for human beings. 

Thus, this study utilizes \citet{lynott2020lancaster}'s sensorimotor norms
 which encompass metrics of sensorimotor strengths (ranging from 0 to 5) of 39,707 concepts spanning six perceptual domains: touch, hearing, smell, taste, vision, and interoception, as well as five action effectors: mouth/throat, hand/arm, foot/leg, head (barring mouth/throat), and torso. To exploit this wealth of data, SensoryT5 is proposed to construct the sensorimotor vectors from these norms and to seamlessly embed them into the T5's decoder mechanism via an auxiliary attention layer. Positioned after the decoders, this sensory-centric attention layer is synergized with the decoder's output, producing an enriched representation brimming with sensory knowledge for words in text. Thus, SensoryT5 is adapted at simultaneously discerning contextual cues and sensory knowledge, allowing for a potent alignment of sensory nuances with contextual intelligence. This integration augments the model's efficacy in the fine-grained emotion classification.
 


\section{Our proposed SensoryT5 model}
\begin{figure*}[!htb] 
\centering 
\includegraphics[width=0.9\textwidth]{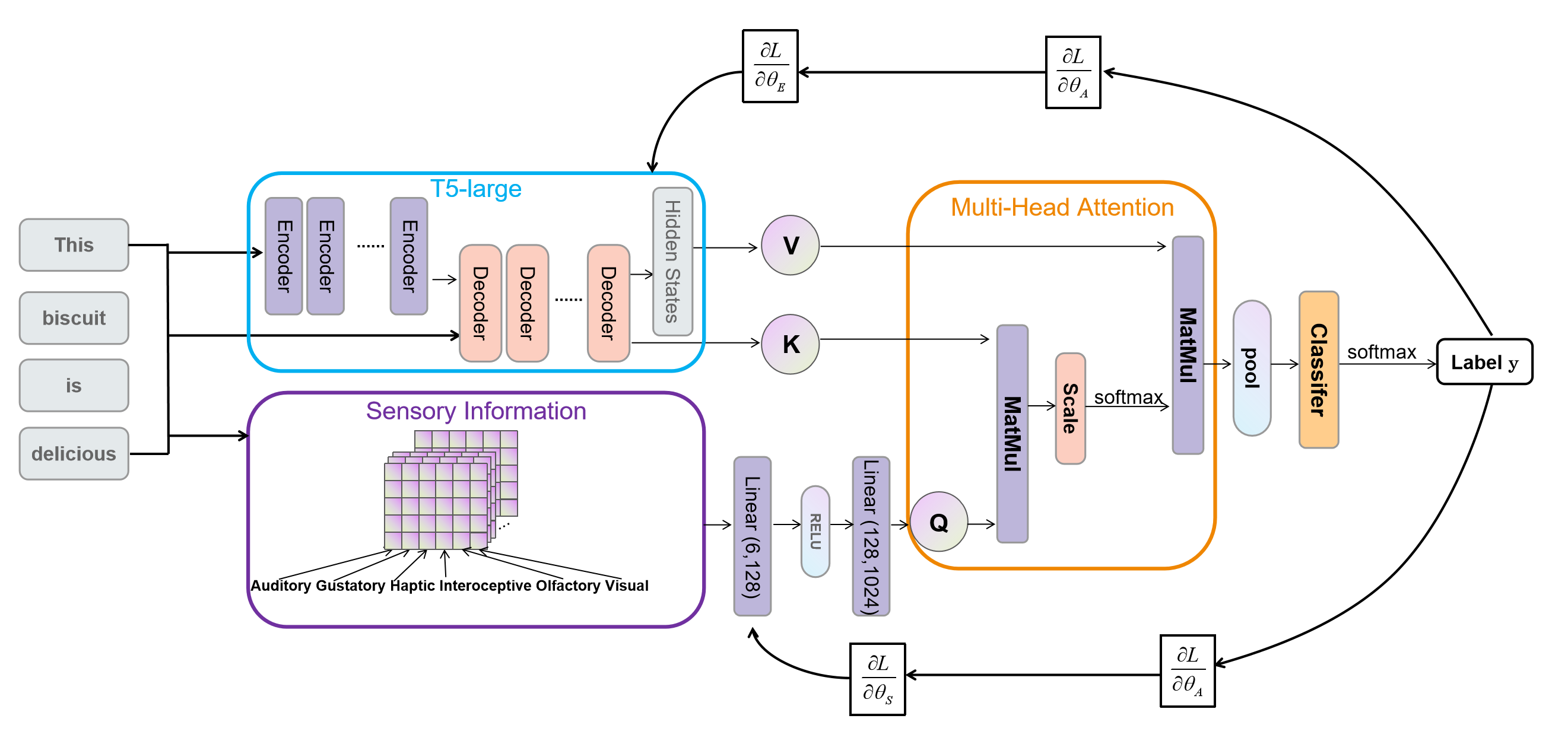} 
\caption{An overview of SensoryT5. Blue box shows a T5 process of deep learning, while purple box describing sensory information is quantified and passed into the T5.} 
\label{Fig:1} 
\end{figure*}
In this section, we elaborate how our SensoryT5 model incorporates the sensory knowledge into the neural emotion classification framework. Specifically, sensory knowledge is infused into the T5 using an adapter approach built upon attention mechanisms. Moreover, the contextual and sensory information learning branches are amalgamated within a unified loss function to facilitate joint training. The overarching structure is depicted in Figure \ref{Fig:1}. 

\subsection{Preliminaries}
\label{sec:3.1}
Despite the relatively large size of the Lancaster Sensorimotor Norms, there are still many out-of-vocabulary words. Following the method proposed by \citet{li2017inferring}, we use a word embedding model to regressively predict the sensory values of unknown words, aiming to obtain sensory values for out-of-vocabulary words.\\\textit{Inputs and outputs} The objective of emotion analysis is to determine and categorize opinions for a piece of texts following a defined label schema. Let $D$ denote a collection of documents for emotion classification. Each document $d \in D$ is first tokenized into a word sequence with maximum length $n$, then the word embeddings $w_i$ of these sequence are jointly employed to represent the document $d = {w_1, w_2, . . . , w_i, . . . , w_n}(i \in 1, 2, . . . , n)$.
\subsection{The core attention mechanism in T5}
The word embeddings of these sequence $d = {w_1, w_2, . . . , w_i, . . . , w_n}(i \in 1, 2, . . . , n)$ first enters the T5. Each layer of the encoder and decoder has a series of multi-head attention units. The multi-head attention mechanism for the final decoder layer can be represented using the following equation:

\begin{equation}
\begin{aligned}
V_d & = \text{MultiHead}(Q_0, K_0, V_0) \\
&  = [\text{head}_1, \text{head}_2, ..., \text{head}_i]W_O
\end{aligned}
\end{equation}

Where each head is computed as:

\begin{equation}
\begin{aligned}
\text{head}_i & = \text{Attention}(Q_0W_i^Q, K_0W_i^K, V_0W_i^V) \\
& = \text{softmax}\left(\frac{(Q_0W_i^Q)(K_0W_i^K)^T}{\sqrt{d_k}}\right)V_0W_i^V
\end{aligned}
\end{equation}

$W_i^Q$, $W_i^K$, and $W_i^V$ are weight matrices that are learned during the training process. They are used to project the input queries ($Q$), keys ($K$), and values ($V$) to different sub-spaces. \( Q_0 \), \( K_0 \),  and \( V_0 \) are derived from the output of the penultimate decoder layer. Additionally, following the common practice for text classification with T5, we employ a zero-padding vector as the sole input for the decoder.
 The result \( V_d \) is the output of the T5 decoder, imbued with context-aware attention. Both \( V_d \) and \( K_0 \) will be utilized in section \ref{sec:3.4} for integration with sensory knowledge.

\subsection{Sensory information transformation for T5 integration}
We project the Lancaster Sensorimotor Norms into a sensory word vector space. Each word is linked with a six-dimensional vector representing sensory scores across six perceptual modalities (auditory, gustatory, haptic, interoceptive, olfactory and visual dimensions). For a word $w$, its sensory vector is denoted as $s(w) = [s_1, s_2, ..., s_6]$.

To enable effective integration into the T5-large, we use two linear transformations followed by a ReLU activation function to map the sensory vectors to the same dimension as the T5-large's word embeddings. Given a T5-large model with an embedding dimension of $1024$ , the transformation process can be formally described as:

\begin{equation}
\mathbf{h}_1 = \text{ReLU}(\mathbf{W}_1 s(w) + \mathbf{b}_1)
\end{equation}

\begin{equation}
s'(w) = \mathbf{W}_2 \mathbf{h}_1 + \mathbf{b}_2
\end{equation}

where $\mathbf{W}_1: \mathbf{R}^6 \rightarrow \mathbf{R}^{128}$ and $\mathbf{W}_2: \mathbf{R}^{128} \rightarrow \mathbf{R}^{1024}$ are two linear transformation matrices and $\mathbf{b}_1$, $\mathbf{b}_2$ are the respective bias terms.The shapes of the two weight matrices $W_1$ and $W_2$ are respectively $(6,128)$ and $(128,1024)$. The output $h_1$ of the first linear layer is a vector of shape $(1,128)$, and the output $s'(w)$ of the second linear layer is a vector of shape $(1,1024)$. After the transformation, the sensory vector $s'(w)$ is projected into the same semantic space as the features generated by T5-large. The output vector $s'(w)$, with $V_d$ and $K_d$ from the T5, will be applied in section \ref{sec:3.4} for infusing sensory knowledge into T5.

\subsection{Sensory attention mechanism in SensoryT5}
\label{sec:3.4}
The sensory vector $s'(w)$ generated by the sensory vector transformation is used as the queries in the attention mechanism of the sensory adapter, substituting the query vector $Q$ in the T5. The sensory adapter performs the attention calculation as follows:

\begin{equation}
\begin{aligned}
A_d & = \text{MultiHead}(s'(w), K_0, V_d) \\
&  = [a_1, a_2, ..., a_i]W_d
\end{aligned}
\end{equation}
where each head is computed as:

\begin{equation}
\begin{aligned}
a_i & = \text{Attention}(s'(w)W_i^Q, K_0W_i^K, V_dW_i^V) \\
& = \text{Softmax}\left(\frac{(s'(w)W_i^Q) (K_0W_i^K)^T}{\sqrt{d_k}}\right)V_dW_i^V
\end{aligned}
\end{equation}

Once the output $A_d = {a_1, a_2, ..., a_n}$ of the sensory adapter is obtained, we apply dropout and pooling operations to form a final representation $P_d$, which is then used as the input to the classification layer.

\begin{equation}
P_d = \text{Dropout}(\text{Pool}(A_d))
\end{equation}
The pooled representation $P_d$ is then fed into the classifier of the T5.
\begin{equation}
C_d = \text{Softmax}(Linear(\text{Dropout}(P_d)))
\end{equation}

$C_d$ is a probability distribution vector. The class with the highest probability is selected as the predicted label, denoted as $y$.

The first step of the back-propagation process involves computing the gradient of the loss function with respect to the parameters of sensory attention adapter. $\Theta_{\text{A}}$ represents the parameters of the sensory attention layer, and $A_d$ represents the output of the sensory T5. The computed gradient is used to update the parameters of the attention layer, enhancing its capacity to integrate sensory information into the T5 model. This is computed as follows:

\begin{equation}
\frac{\partial \mathcal{L}}{\partial \Theta_{\text{A}}} = \frac{\partial \mathcal{L}}{\partial A_d} \cdot \frac{\partial A_d}{\partial \Theta_{\text{A}}}
\end{equation}

After the gradients for the sensory attention mechanism have been computed, we then compute the gradients for the parameters of the final layer of the T5, denoted as $\Theta_{\text{E}}$.

\begin{equation}
\frac{\partial \mathcal{L}}{\partial \Theta_{\text{E}}} = \frac{\partial \mathcal{L}}{\partial V_d} \cdot \frac{\partial V_d}{\partial \Theta_{\text{E}}}
\end{equation}

Finally, the gradients for the sensory information transformation, denoted as $\Theta_{\text{S}}$, are computed as follows:

\begin{equation}
\frac{\partial \mathcal{L}}{\partial \Theta_{\text{S}}} = \frac{\partial \mathcal{L}}{\partial s'(w)} \cdot \frac{\partial s'(w)}{\partial \Theta_{\text{S}}}
\end{equation}
Here, $\Theta_{\text{S}}$ represents the parameters of the sensory information transformation component, which includes the weights and biases of the two linear layers, and $s'(w)$ represents the output of this component. The calculated gradient is used to update the parameters of the sensory information transformation to improve its ability to capture and model sensory information.
Through these calculations, we are able to update the parameters of the sensory attention mechanism, the T5, and the sensory information transformation component.

\section{Experimental evaluation}

\subsection{Datasets}
We have selected four benchmark datasets of varying sizes to encompass a variety of classification tasks: Empathetic Dialogues (ED) \cite{rashkin-etal-2019-towards}, GoEmotions (GE) \cite{demszky-etal-2020-goemotions}, ISEAR \cite{scherer1994evidence} and EmoInt \cite{mohammad-bravo-marquez-2017-wassa}. For the GE dataset, we exclusively utilize samples with a single label and omit those that are neutral to maintain an equitable comparison with prior studies \cite{suresh-ong-2021-negatives, chen2023label}. Table~\ref{tab:dataset} presents a summary of key statistics for these datasets. 
Our evaluation utilizes two widely recognized performance metrics: accuracy and the F1 score, in line with state-of-the-art studies.

\begin{table}[htb]
\label{tab:dataset}
\centering

\begin{tabular}{l|c|c|c|c}
\hline
Dataset & $N_{\text{train}}$ & $N_{\text{test}}$ & L & C \\
\hline
ED & 19,533 & 2,547 & 18  & 32 \\
GE   & 23,485 & 2,984& 12  & 27 \\
ISEAR   & 4,599 & 1,534 & 22 & 7 \\
EmoInt & 3,612 & 3,141 & 16 & 4 \\

\hline
\end{tabular}
\caption{Statistics of the four benchmark datasets. In the table, "$N_{\text{train}}$" and "$N_{\text{test}}$" respectively represent the number of instances in the training and testing sets. "L" stands for the average text length within the dataset, and "C" indicates the number of classes/categories.}
\label{tab:dataset}
\end{table}

\subsection{Sensory knowledge}

\begin{figure*}[!htb] 
\centering 

\includegraphics[width=0.9\textwidth]{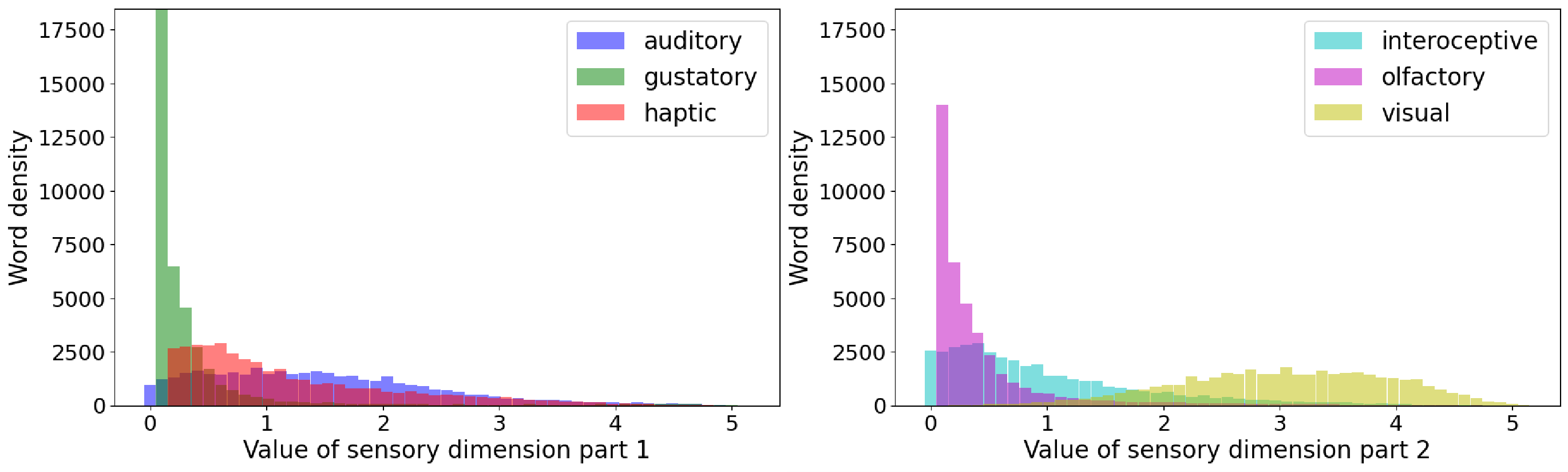} 
\caption{Histograms showing the distribution six sensory values over words. X-axis shows the value in an sensory dimension, while y-axis displays the word density.} 
\label{lancaana} 
\end{figure*}

Before conducting the emotion analysis experiments, we conducted a preliminary analysis of our sensory lexicons from the perspective of sensory perception value distribution. Figure \ref{lancaana} displays histograms of the six sensory measures across all words within our model. Notably, the distributions for these measures are quite unbalanced. Gustatory and olfactory measures predominantly demonstrate a left-skewed distribution, with most values ranging between 0 and 1. This suggests that these two sensory perceptions are less frequently represented in the textual context. Thus, it might be challenging to represent gustatory and olfactory perceptions from text.

In contrast, auditory and visual measures show a relatively uniform distribution. The auditory measure is evenly distributed between 0 and 2.5, while the visual measure ranges between 2 and 4.5. These  distributions indicate a higher sensitivity of auditory and visual knowledge to textual information, which suggests that auditory and visual senses may play a significant role within sensory models. 


Lastly, haptic and interoceptive measures exhibit similar trends, declining from about 2500 to 0 as the values increase from 0 to 5. The decline in the presence of 
haptic and interoceptive knowledge across the general textual context might suggest that they are less informative sensory dimensions in the majority of cases.




As discussed in section \ref{sec:3.1}, the Lancaster Sensorimotor Norms dataset is subject to size limitations, resulting in a significant number of unknown words for which corresponding sensory values are unavailable. To address this challenge, we adopted the method proposed by \citet{li2017inferring} for predicting sensory values of unknown words through embedding techniques. In our experiments, we utilized both the T5 embedding and the GloVe embedding \citep{pennington2014glove} for this prediction task.

To assess the accuracy of our predictions, we randomly selected 10\% of the Lancaster Sensorimotor Norms dataset as a validation set and applied the Root Mean Square Error (RMSE) as the evaluation metric. The experimental results are presented in Table \ref{tab:extend}. The results demonstrate that GloVe outperforms T5 Embedding in predicting each sensory dimension. To preserve the original features of the Lancaster dataset to a minimal extent, we opted for a smaller version of GloVe with 400,000 data points and 200 dimensions. Following augmentation, our sensory vocabulary size reached 407,572\footnote{We will release the sensory vector after this paper is accepted}.

\begin{table}[ht]
\centering

\begin{tabular}{c|c|c}
\hline
 \textbf{Sensory Name}& \textbf{T5 Embedding} & \textbf{GloVe}\\

\hline
Auditory & 0.949 & \textbf{0.803}\\
Gustatory & 0.632 & \textbf{0.534}\\
Haptic & 0.893 & \textbf{0.698}\\
Interoceptive & 0.831 &\textbf{ 0.662}\\
Olfactory & 0.572 & \textbf{0.501}\\
Visual & 0.842 & \textbf{0.743}\\
\hline
Total & 0.798 & \textbf{0.665}\\

\hline
\end{tabular}
\caption{Comparison of prediction accuracy between T5 Embedding and GloVe techniques on different sensory dimensions, as measured by RMSE values. Lower scores indicate higher accuracy in the prediction of sensory values.}
\label{tab:extend}
\end{table}

For validating our augmentation, we evaluated the coverage rates of sensory word vectors before and after augmentation across all datasets we employed, as detailed in Table \ref{wordrate}. As evident from the augmentation results, the coverage range significantly expands in comparison to the original data across all datasets. This underscores the enhanced impact of integrating sensory information into the model on the results.

\begin{table}[ht]
\centering

\begin{tabular}{c|c|c}
\hline
 \textbf{Datasets}& \textbf{Lancaster \%} & \textbf{Exten-Lancaster \%}\\

\hline
ED & 58.23 & 91.78\\
GE & 46.85 & 83.91\\
ISEAR & 54.62 &78.97\\
EmoInt & 29.65 & 46.21\\

\hline
\end{tabular}
\caption{ Word coverage of Lancaster Sensorimotor Norms before and after expansion using regression prediction.}
\label{wordrate}
\end{table}
\subsection{Experiment settings and Baselines}
We compare the proposed SensoryT5 primarily with two group of strong baselines:

\textbf{PLMs.} We
compared against BERT \cite{devlin-etal-2019-bert}, RoBERTa \cite{liu2019roberta}, XLNet \cite{yang2019xlnet} and T5 \cite{raffel2020exploring}. The advent of PLMs has marked a significant improvement across a multitude of tasks in the realm of natural language processing, including text classification. This leap in performance is largely due to the deep and nuanced semantic representations these models extract from the text, facilitating a more profound understanding and interpretation of linguistic content.

\textbf{Label Embedding-aware models.} \citet{suresh-ong-2021-negatives} introduced a concept called label-aware contrastive loss (LCL). This technique uniquely assigns varying weights to each negative sample. Importantly, pairs that are more easily confounded have a higher impact on the objective function, enhancing outcomes in fine-grained text classification scenarios.
\citet{chen2023label} proposed HypEmo, a framework enhancing fine-grained emotion classification by utilizing hyperbolic space for label embedding. This model integrates hyperbolic and Euclidean geometries to discern subtle nuances among labels effectively. 

These two models, LCL and HypEmo, stand as the most potent in the realm of fine-grained emotion classification, delivering unparalleled results due to their innovative handling of nuanced label distinctions and hierarchical intricacies.

\textbf{Implementation Details.}\footnote{We will release the open-source code after this paper is accepted}
During training, we applied the Adam optimizer in Euclidean space. We set the learning rate at a consistent $10^{-4}$, maintaining a balance between rapid adaptation and the stability of learning, reducing the likelihood of oscillation or divergence. 


\begin{table*}[!htb]
\centering

\small 
\begin{tabular}{lcccc|cccc}
\hline
 \multicolumn{1}{c}{} & \multicolumn{2}{c}{Empathetic Dialogue} & \multicolumn{2}{c|}{GoEmotions} & \multicolumn{2}{c}{ISEAR} & \multicolumn{2}{c}{EmoInt} \\

 & ACC & F1 & ACC & F1 & ACC & F1 & ACC & F1  \\
\hline

\textasteriskcentered $ \text{BERT}_{\text{large}} $

 & 0.557 & 0.551 & 0.642 &0.637 & 0.677 &0.679  &  0.848 &0.848   \\

\textasteriskcentered $ \text{RoBERTa}_{\text{large}} $
  & 0.596 &0.590  &  0.652& 0.644&0.723  &0.720  &0.865   &0.865   \\
\textasteriskcentered  $ \text{XLNet}_{\text{large}} $ &0.599  &0.592  & 0.641 &0.568 & 0.711 & 0.711 &0.845   &0.845   \\
\textasteriskcentered $ \text{T5}_{\text{large}} $ & 0.609 & 0.604 & 0.661 & 0.657 & 0.717& 0.717& 0.863 &  0.863 \\
 \hline

 \textdagger LCL & 0.601 & 0.591 & 0.655 & 0.648 & 0.724 &0.724 & 0.866 & 0.866  \\

\S HypEmo & 0.596 & 0.610 & 0.654 & 0.663 &0.707  & 0.712 &0.846  &0.846   \\
\hline
\textasteriskcentered  SensoryT5 & \textbf{0.618} & \textbf{0.615} & \textbf{0.674} & \textbf{0.670}& \textbf{0.726} & \textbf{0.724} & \textbf{0.875}  & \textbf{0.875}  \\

\hline
\end{tabular}
\caption{Evaluation on fine-grained emotion classification, the result with the best performance are highlighted in bold. Data marked with \textdagger are from \citep{suresh2021not}, \S  from \citep{chen2023label}, and \textasteriskcentered represents our own results. Note: In \S, results from missing datasets (ISEAR and EmoInt) were supplemented by our experiments.}
\label{maintable}
\end{table*}

\subsection{Baseline comparison}
To demonstrate the effectiveness of SensoryT5, we embarked on a comprehensive set of comparative experiments, analyzing its performance in emotion classification tasks. The comparison is shown in Table \ref{maintable}. Firstly, we
compare SensoryT5 with PLMs. SensoryT5 registers an impressive enhancement over T5's performance, the best of the PLM contenders. For instance, SensoryT5 exhibits an increase in accuracy by 0.9\% for Empathetic Dialogues and 1.3\% for GoEmotions, showcasing its finesse in handling diverse emotional contexts. This upward trend continues with ISEAR and EmoInt datasets, where SensoryT5 improves by 0.9\% and 1.2\%, respectively, over T5.

Secondly, we compare with label-aware system. These two models, LCL and HypEmo, stand as the most potent in the realm of fine-grained emotion classification. LCL outperforms T5 in the ISEAR and EmoInt datasets, while the other datasets under the label-aware system category do not compete favorably with T5. This comparative analysis is critical, considering that LCL utilizes a synonym substitution technique to effectively double its dataset size. Such an expansion contributes significantly to its enhanced performance metrics.
In our experiments, we strictly adhered to using original samples without resorting to any form of data augmentation techniques. Despite this, SensoryT5 surpasses LCL by 0.2\% and 0.9\% in accuracy on the ISEAR and EmoInt datasets, respectively. This margin of improvement, although seemingly nominal, is quite significant in the context of these tasks. It underscores the efficacy of our proposed method of infusing sensory perceptions into the model.

In summary, compared to previous studies, we have achieved superior results without the necessity for additional data, marking the current pinnacle in this field. This accomplishment underscores the effectiveness of SensoryT5.

\subsection{Ablation studies}

\begin{figure*}[!htb] 
\centering 

\includegraphics[width=0.9\textwidth, height=0.18\textwidth]{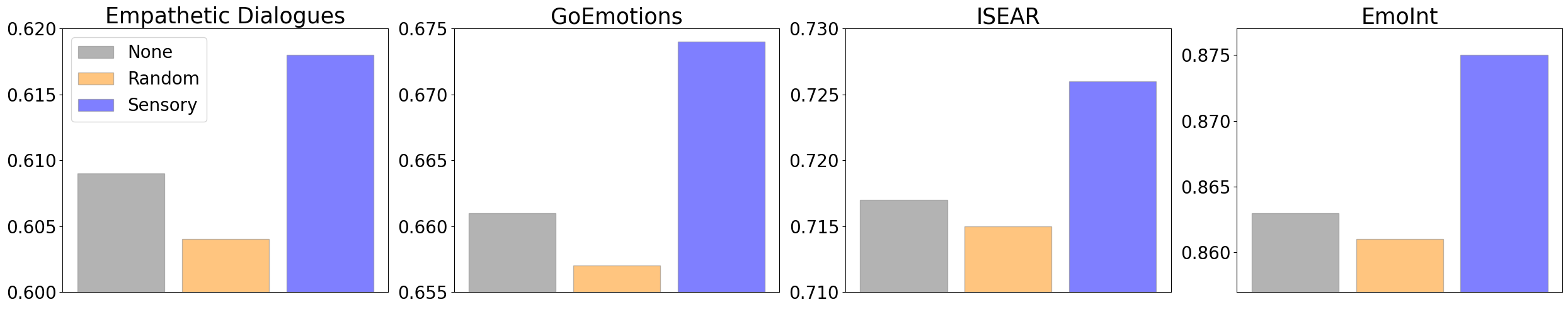} 
\caption{Ablation Study Results. Performance of T5 (None), Random SensoryT5 (with sensory values randomly assigned), and SensoryT5 across four datasets, evaluated using accuracy as the metric.} 
\label{aba} 
\end{figure*}
\begin{figure*}[!htb] 
\centering 

\includegraphics[width=0.9\textwidth, height=0.3\textwidth]{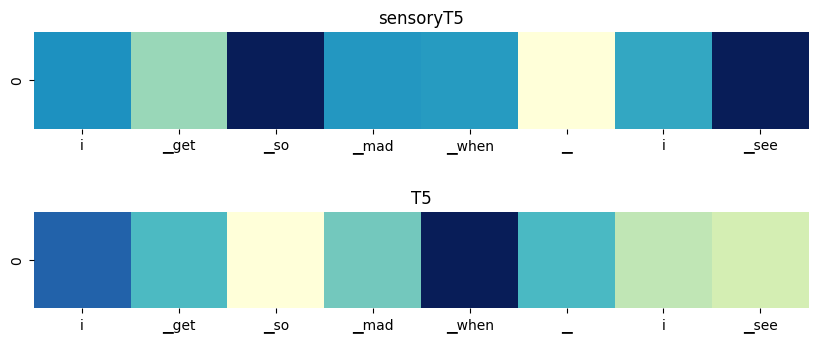}

\caption{The heat values of the final sensory layer in SensoryT5 and the encoder layer in T5 for the sentence 'I get so mad when I see or hear about kids getting bullied...' sourced from the Empathetic Dialogues training dataset.} 
\label{case} 
\end{figure*}
In our efforts to understand the contributions of different components within the SensoryT5 model, we conducted ablation studies, a critical methodological step in assessing the impact of our novel sensory integration. These studies were also carried out on four datasets. The ablation tests were structured around three primary configurations:

\textbf{SensoryT5:} Our complete model infusing sensory information.

\textbf{Random SensoryT5:} A variant of our model where the sensory values were substituted with random numbers ranging from 0 to 5, maintaining the same distribution of sensory scores but eliminating their meaningful association with the data.

\textbf{T5 (None):} The baseline model without any sensory information, representing the standard PLM approach in fine-grained emotion classification tasks.

The result is shown in Figure \ref{aba}. While the SensoryT5 model exhibited the highest performance in terms of accuracy across all datasets, the Random SensoryT5 configuration yielded lower results than even the T5. This decrement in performance was especially pronounced on the more complex datasets, Empathetic Dialogues and GoEmotions.

The degradation in performance with random sensory values underscores the importance of meaningful sensory integration. It is not merely the presence of additional numerical data that enhances the SensoryT5 model's performance, but rather the contextually relevant and accurately associated sensory information that it brings to the emotion classification task.

Furthermore, the fact that the Random SensoryT5 underperformed compared to the T5 indicates that arbitrarily added sensory information could introduce noise into the model, disrupting its ability to correctly interpret and classify emotional content. This revelation is significant, affirming that the strategic integration of sensory data is crucial, and haphazard integration could be counterproductive.

In summary, these ablation studies have confirmed the value of our sensory information layer, as evidenced by the performance drop when this layer is randomized or removed. This reinforces our assertion that the SensoryT5's strength lies in its ability to simulate a more human-like understanding of textual data, resonating with how humans perceive emotions through a sensory lens.

\subsection{Case study}

We conducted a focused case study on the SensoryT5 model using a sentence from the Empathetic Dialogues dataset: "I get so mad when I see or hear about kids getting bullied..."
In Figure \ref{case}, attention heatmaps display the model's focus during processing. The SensoryT5 heatmap shows the aggregate attention for each token in the sensory layer, while the T5 section compiles attention weights across all encoder layers, subsequently averaging them to reveal the model's overall focus. The SensoryT5 model exhibited intensified attention on the emotionally significant phrase "so mad," highlighting its ability to detect crucial emotional nuances. In contrast, the standard T5's attention was more distributed, less focused on the emotional pivot. This micro-level analysis reveals SensoryT5's superior capability in recognizing emotional cues. Such insights substantiate the efficacy of integrating sensory awareness into language models for improved emotional discernment.

In summary, our extensive evaluations and comparative studies highlight the superior performance of SensoryT5 over other PLMs based emotion classification models, including the T5. When benchmarked against the state-of-the-art methods, SensoryT5 notably surpassed them, establishing a new standard in the field. Further, our ablation studies convincingly demonstrate that the effectiveness of SensoryT5 is attributed more to its integration of sensory perception than to structural enhancements. This assertion is corroborated by our detailed case studies, which offer a microscopic view into the instances where SensoryT5's unique capabilities are distinctly evident. Collectively, these findings underscore a breakthrough performance of SensoryT5 in the realm of fine-grained emotion classification. Importantly, it signifies a successful adaptation within the shift towards incorporating neuro-cognitive data in NLP, validating the premise that a deeper convergence between sensory data and language modeling leads to a more profound understanding of emotional nuances.
\section{Conclusion}
In this paper, we propose the SensoryT5 model designed for the fine-grained emotion classification. This framework harnesses sensory knowledge, aiming to boost the prowess of transformers in pinpointing nuanced emotional subtleties. By integrating sensory knowledge into T5 through attention mechanisms, the model concurrently evaluates sensory cues alongside contextual hallmarks. Crucially, SensoryT5 exhibits exceptional adaptability and precision, making it a formidable tool for tasks in Fine-grained Emotion Classification, including configurations like 32-class, 27-class, 7-class, and 4-class delineations. Moreover, SensoryT5 serves as a conduit between sensory perception and emotional understanding, embodying the recent paradigm shift in NLP towards a more neuro-cognitive approach. It acknowledges and capitalizes on the intrinsic relationship between our sensory experiences and our emotional responses, a connection well-documented in neuro-cognitive science but often under-explored in computational fields. By interpreting sensory lexicon through advanced representation learning, SensoryT5 decodes the implicit emotional undertones conveyed, mirroring the human ability to associate sensory experiences with specific emotional states. In recognizing the entwined nature of cognition, sensation and emotive expression, SensoryT5 not only contributes to but also encourages the continuation of interdisciplinary research efforts. It stands as testament to the potential of a more nuanced and integrative approach in NLP, where understanding language transcends the boundaries of words and grammar, delving into the very experiences and perceptions that shape human emotionality.

\section*{Limitations}
In our work, we utilized GloVe and T5 embeddings to predict sensory values for unknown words using a regression method. This approach learns only from static values. To derive static T5 embeddings, we passed all tokens sequentially through the T5 embedding layer, obtaining a static embedding for each token. This process, however, leads to a limitation: it compromises the original dynamic context-embedding capabilities of T5. In T5 embeddings, different embeddings are obtained based on the different contexts. We intended to learn from these transformer embeddings and then predict.
Additionally, when compared to current state-of-the-art models in emotion classification, such as the label embedding-aware HypEmo and LCL, SensoryT5 exhibits certain inadequacies, particularly in terms of interpretability. Both HypEmo and LCL not only surpass SensoryT5 in explaining their decision-making processes but also do so with fewer parameters. These models, by leveraging sophisticated label-aware embedding strategies, provide insights into the nuanced relationships and hierarchies among labels, something that SensoryT5, with its reliance on static values, struggles to achieve. This gap highlights a significant area for improvement in SensoryT5, suggesting the need for an advanced approach that maintains the richness of context-sensitive embeddings while enhancing the model's overall interpretability and efficiency.

\nocite{*}
\section{Bibliographical References}\label{sec:reference}

\bibliographystyle{lrec-coling2024-natbib}
\bibliography{lrec-coling2024-example}

\end{document}